# Personalized Influence Estimation Technique

[a*] Kumarjit Pathak, [b*]Jitin Kapila, [c] Aasheesh Barvey

**Abstract:** Customer Satisfaction is the most important factors in the industry irrespective of domain. Key Driver Analysis is a common practice in data science to help the business to evaluate the same. Understanding key features, which influence the outcome or dependent feature, is highly important in statistical model building. This helps to eliminate not so important factors from the model to minimize noise coming from the features, which does not contribute significantly enough to explain the behavior of the dependent feature, which we want to predict.

Personalized Influence Estimation is a technique introduced in this paper, which can estimate key factor influence for individual observations, which contribute most for each observation's behavior pattern based on the dependent class/estimate. Observations can come from multiple business problem i.e. customers related to satisfaction study, customer related to Fraud Detection, network devices for Fault detection etc. It is highly important to understand the cause of issue at each observation level to take appropriate Individualized action at customer level or device level etc. This technique is based on joint behavior of the feature dimension for the specific observation, and relative importance of the feature to estimate impact.

The technique mentioned in this paper is aimed to help organizations to understand each respondent's or observation's individual key contributing factor of Influence. Result of the experiment is really encouraging and able to justify key reasons for churn for majority of the sample appropriately.

***Index Terms***— **Key Driver, Variable Importance, Attribute Impact, Personalized Influence**

## I. INTRODUCTION

In current market situation understanding customer based on individual prospective is very important. Individualization of attention and treatment are the key for customer retention and winning criteria.

Taking actions based on overall key drivers is an outcome of a data science algorithm which can assist to detect overall importance of the features which helps business to derive overall strategy to increase satisfaction of customer base. However, where dynamic personalized business intervention is the ask; overall key drivers coming out from any algorithm does not suffice to point out key concern area or high impact touch point for specifically for any one of the customer.

In response to this problem we propose a methodology to detect key influence by each customer or observation in the data.

## II. LITERATURE REVIEW AND RELATED WORK

There are quite a lot of research has happened on the "Key Driver Analysis"

Marco Tulio et al[11] has introduced t Local Interpretable Model-agnostic Explanations (LIME). The overall goal of LIME is to identify an interpretable model over the interpretable representation that is locally faithful to the classifier overall Key drivers based on different methodology.

However, this approach come with a caveat that all observations may not be impacted by all the derived important features or explanatory variables. It is important

**Algorithm 1** Sparse Linear Explanations using LIME

**Require:** Classifier $f$, Number of samples $N$
**Require:** Instance $x$, and its interpretable version $x'$
**Require:** Similarity kernel $\pi_x$, Length of explanation $K$
$\quad \mathcal{Z} \leftarrow \{\}$
$\quad$ for $i \in \{1, 2, 3, ..., N\}$ do
$\quad\quad z_i' \leftarrow sample\_around(x')$
$\quad\quad \mathcal{Z} \leftarrow \mathcal{Z} \cup \langle z_i', f(z_i), \pi_x(z_i) \rangle$
$\quad$ end for
$\quad w \leftarrow$ K-Lasso$(\mathcal{Z}, K)$ $\quad \triangleright$ with $z_i'$ as features, $f(z)$ as target
$\quad$ return $w$

Fig. 1. Sparse Linear Explanation algorithm by Marco Tulio et al[11].

that each feature can weighted at each observation level to understand for that observation which feature is key factor for personalized attention for any marketing action.

Authors:
[a*] Mr. Kumarjit Pathak, Data Scientist, Working @ Harman Connected Services India Pvt Bengaluru, Karnataka- 560066 (e-mail: Kumarjit.pathak@ outlook.com).
[b*] Mr. Jitin Kapila, Data Scientist, Working @ Zetaglobal, Bengaluru, Karnataka- 560066 (e-mail: Jitin.kapila@ outlook.com).

[c] Mr. Aasheesh Barvey, Data Scientist Working @ Harman Connected Services India Pvt Ltd., Bengaluru, Karnataka- 560066 (e-mail: ashbarvey@ gmail.com).

\* Major and equal contribution
.

**Algorithm 2** Submodular pick (SP) algorithm

**Require:** Instances $X$, Budget $B$
  **for all** $x_i \in X$ **do**
    $\mathcal{W}_i \leftarrow \textbf{explain}(x_i, x'_i)$         ▷ Using Algorithm 1
  **end for**
  **for** $j \in \{1 \ldots d'\}$ **do**
    $I_j \leftarrow \sqrt{\sum_{i=1}^{n} |\mathcal{W}_{ij}|}$   ▷ Compute feature importances
  **end for**
  $V \leftarrow \{\}$
  **while** $|V| < B$ **do**      ▷ Greedy optimization of Eq (4)
    $V \leftarrow V \cup \text{argmax}_i \, c(V \cup \{i\}, \mathcal{W}, I)|$
  **end while**
  **return** $V$

Fig. 2. Submodular pick algorithm by Marco Tulio et al[11]..

Laura Funa et al[4] has showcased Various regression, correlation and cooperative game theory approaches were used to identify the key satisfiers and dissatisfiers. The theoretical and practical advantages of using the Shapley value, Canonical Correlation Analysis and Hierarchical Logistic Regression has been demonstrated and applied to market research. However this approach come with a caveat that all customers or observations may not be impacted by all the features or explanatory variables. Hence individual attention is imperative.

Abdullah Hussein Al-Hashed et al [8] introduces SERVPERF model is utilized to measure service quality from functional aspects. Network quality dimension is newly added to measure service quality from technical aspects.

Kevin Gray et al[6] has show cased the difference between stated and derived importance and highlighted the usefulness of Ridge regression, Shapley value regression, stepwise regression and the Lasso to estimate overall key driver.

Kurt Matzler et al[1] explains a technique "importance–performance analysis (IPA) which widely used analytical technique that yields prescriptions for the management of customer satisfaction. IPA is a two-dimensional grid based on customer-perceived importance of quality attributes and attribute performance.

According to all the mentioned methodology overall key diver identification is addressed. These techniques do not really provide and intuition or methodology how the overall key drivers derived from the data can be tagged to individual response. Our proposed methodology aims at addressing this gap and provide business with a tool to provide scope for personified attention to address each respondent or response.

## III. Pie Formulation

Currently in industry Key Drive is well known term and lot of research has gone to find the intrinsic pattern from a given dataset which can summaries major contributing factors towards the estimation of dependent feature.

Let's assume any supervised machine learning model to predict one dependent feature 'Y' based on 'm' independent continuous or count data type feature(x). Algorithm provides key drivers estimate $\beta_k$ where $1 \le k \le m$ then the statistical model can be represented by a function $f(\beta_k, x_k, c_k)$ with some bias term $c_k$.

For simplicity of representation let's consider:

$$Y = f(\beta_k, x_k)$$

For linear supervised model this equation would look like :

$$Y = \beta_1 * x_1 + \beta_2 * x_2 + \cdots + \beta_k * x_k + \cdots + \beta_m * x_m$$

Where $\{\beta_1, \beta_2, \beta_3, \ldots \beta_m\}$ are the importance parameters coming from linear estimation and $\{x_1, x_2, x_3, \ldots x_m\}$ are the features used for the pattern estimation

Then for every single observation in the data set of 'n' observations, For " $i^{\text{th}}$ " observation most important feature towards the target prediction is the feature with maximum $Pie\_Y_i$ value :

**For linear/tree based models:**

For Non – linear model such Neural network, Support Vector Machine, Random Forest where the model is trained on the particular target class or value (in case of regression problem) we can derive most impactful variable with the following technique.

Here we need to first transform the feature set and observation set

$$\beta_k \leftarrow (\beta_k - \bar{\beta}_k)/\sigma\beta$$

$$x_{ik} \leftarrow (x_{ik} - \bar{x}_k)/\sigma x_k$$

Post this transformation, we find soft- threshold to find relative important factor specifically for the observation contributing towards target class occurrence eg: Fraud, Attrition of customer,

$$PIE \text{ of } Y[i] \text{ is feature } x_k \text{ where}$$

$$Pie\_Y_i = Max_{k>0}^{m}\left\{\frac{\beta_k * x_{ik}}{\sum_{k=1}^{m} \beta_k * x_{ik}}\right\}$$

an also look at top few selected number of Pie_y[i] to explore more than one area of focus.

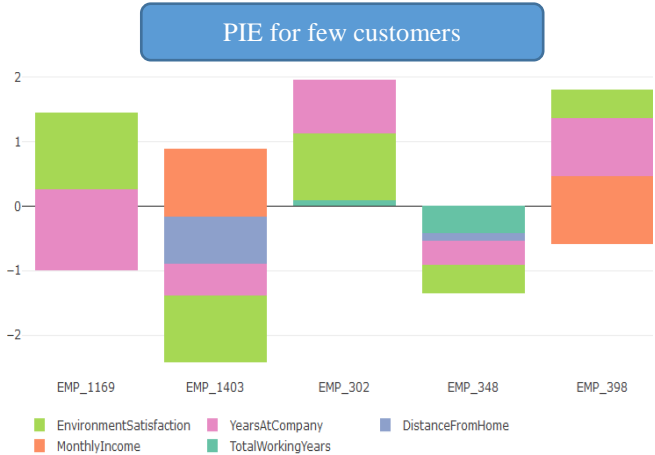

Fig. 1. Example of what is driving the churn for each employee. This is based on IBM published data for churn

---

## Algorithm for $Pie\_Y_i$

Require: Overall feature importance $\beta_k$,

Require: Observations including all the features selected as importnat $x_{ik}$

For k in {1,2,…m}

   $\beta_k \leftarrow (\beta_k - \bar{\beta}_k)/\sigma\beta$

   $\beta k \leftarrow if \ \beta k < 0, then \ \beta k = 0, else, \ \beta k$

For k in {1,2,…..m}

   For i in {1,2,…..n}

      $x_{ik} \leftarrow (x_{ik} - \bar{x}_k)/\sigma x_k$

      $xik \leftarrow if \ xik < 0, then \ xik = 0, else, \ xik$

   End for

End for

$Wi_k \leftarrow \{\}$

$Si_k \leftarrow \{\}$

For i in {1,2,…..n}

If $x_{ik} > 0 \ \& \ 0 < \beta_k$

   $Si_k \leftarrow \sum_{k=1}^{m} \beta_k \ * x_{ik}$

   For k in {1,2,…..m}

      $Wi_k = \frac{\beta_k \ * x_{ik}}{Si_k}$

   End for

$Pie\_Y_i = argmax_k(Wi_k / Si_k)$

End for

---

This approach is not generalized for discrete features.

Idea behind is to cross compare normalized datum weightage with reference to normalized overall feature importance to generate specific key driver for each datum which represents each customer.

Proposed novel method was been tested on different datasets in the organization and it has managed to provide granular key driving factors for each customer uniquely.

However for the purpose of the paper is to display the novelty of the solution without exposing any organization data.

These 3 graphs [fig.2, fig.3 , fig.4] showcase illustrative example. Where Fig 2 show the values of the specific data line. Representing one row of the dataset /each customer. Fig4 provides visual of the most impacting drivers for the selected observation.

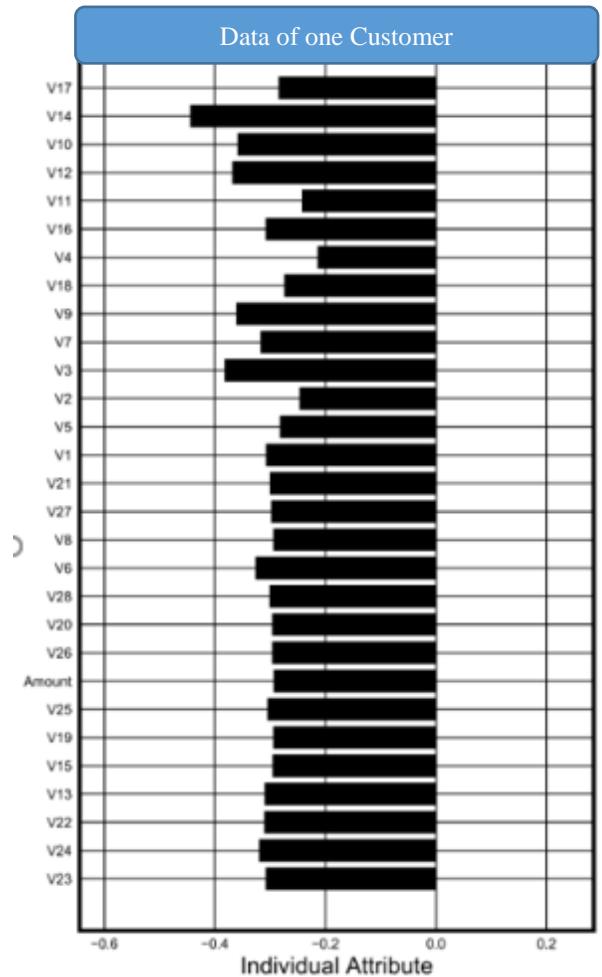

Fig. 2. Example of a datum- representing one customer

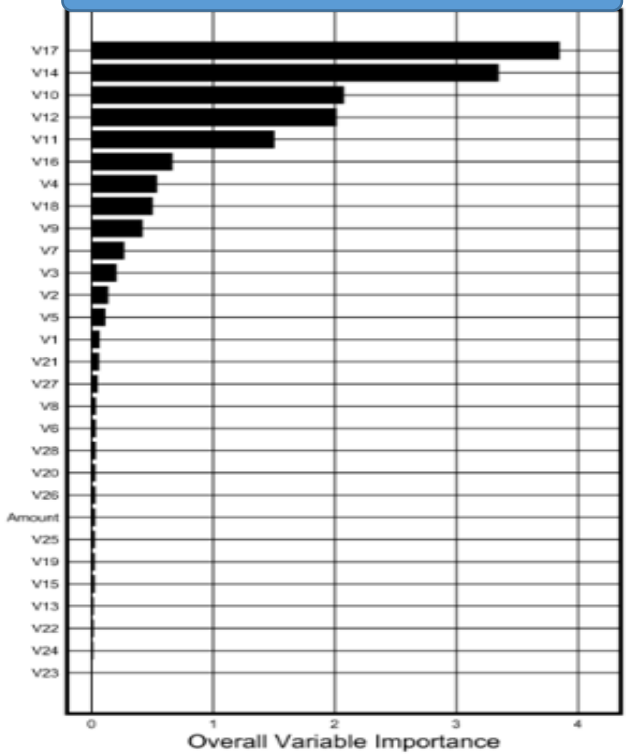

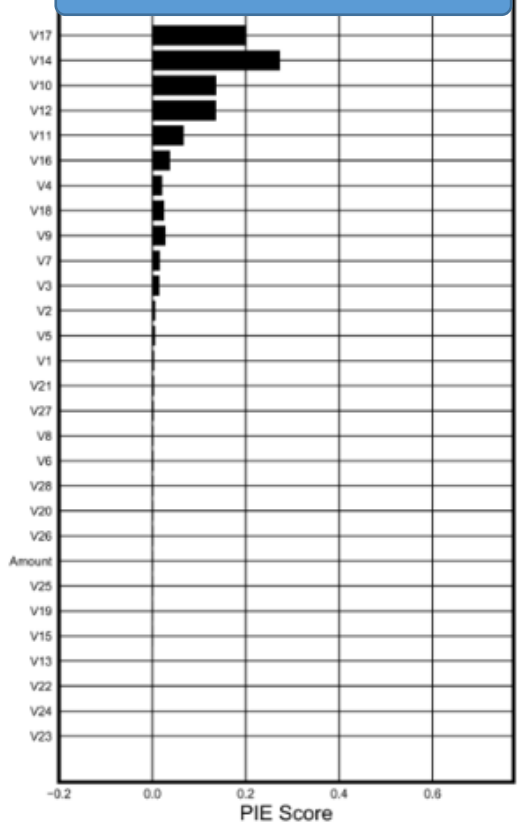

## IV. CONCLUSION

This paper would potentially help any organization to put personal attention with selective engagement to each respondent, based on individual key issue area. Implementing this methodology on customer attrition modelling would help organizations to provide personalized promotion or personalized proactive engagement similarly for fraud detection.

Telecom vertical can use this to find exact cause of equipment failure for each of the equipment separately and provide support intelligence to engineers and reduce cycle time of repair.

## REFERENCES

[1] Kurt Matzlera,*, Franz Bailomb,1, Hans H. Hinterhubera,2, Birgit Renzla,3, Johann Pichlerb,1. "The asymmetric relationship between attribute-level performance and overall customer satisfaction: a reconsideration of the importance–performance analysis." Department of General and Tourism Management, University of Innsbruck, Universita¨tsstrasse 15, A-6020 Innsbruck, Austria "Innovative Management Partner (IMP), Rennweg 23, A-6020 Innsbruck, Austria. March 20034.

[2] Michael Conklin, Ken Powaga and , Stan Lipovetsky "Customer Satisfaction Analysis: Identification of Key Drivers" European Journal of Operational Research, 2004, 154/3, 819-827.

[3] Wei-JawDeng and WenPei, "Fuzzy neural based importance-performance analysis for determining critical service attributes" March 2009.

[4] Laura Funa " Customer Satisfaction Analysis" Dec 2017, published on DiVA - Academic Archive.

[5] Sarah Marley - "Key Driver Analysis" Available: URL https://select-statistics.co.uk/blog/key-driver-analysis/.

[6] Kelvin Gray - "Best practices for key driver analysis" Feb-2014, published on QUIRK'S Media.

[7] Sumit Agarwal, Dr. Deepak Singh, Prof. K S Thakur, "The Impact of Service Quality Dimensions on Customer" , Pacific Business Review International, Volume 6, Issue 1, July 2013.

[8] Abdullah Hussein Al-Hashedi1,*, Sanad Ahmed Abkar- "The Impact of Service Quality Dimensions on Customer Satisfaction in Telecom Mobile Companies in Yemen" [American Journal of Economics 2017, 7(4): 186-193].

[9] Jeff Sauro – "10 THINGS TO KNOW ABOUT A KEY DRIVER ANALYSIS" , October 18, 2016.

[10] Calvo-Porral - "Smooth operators? Drivers of customer satisfaction and switching behavior in virtual and traditional mobile services" September- 2015 Science Direct.

[11] Marco Tulio Ribeiro, Sameer Singh, Carlos Guestrin ""Why Should I Trust You?": Explaining the Predictions of Any Classifier" ArXiv Submitted on 16 Feb 2016 (v1), last revised 9 Aug 2016 (this version, v3))

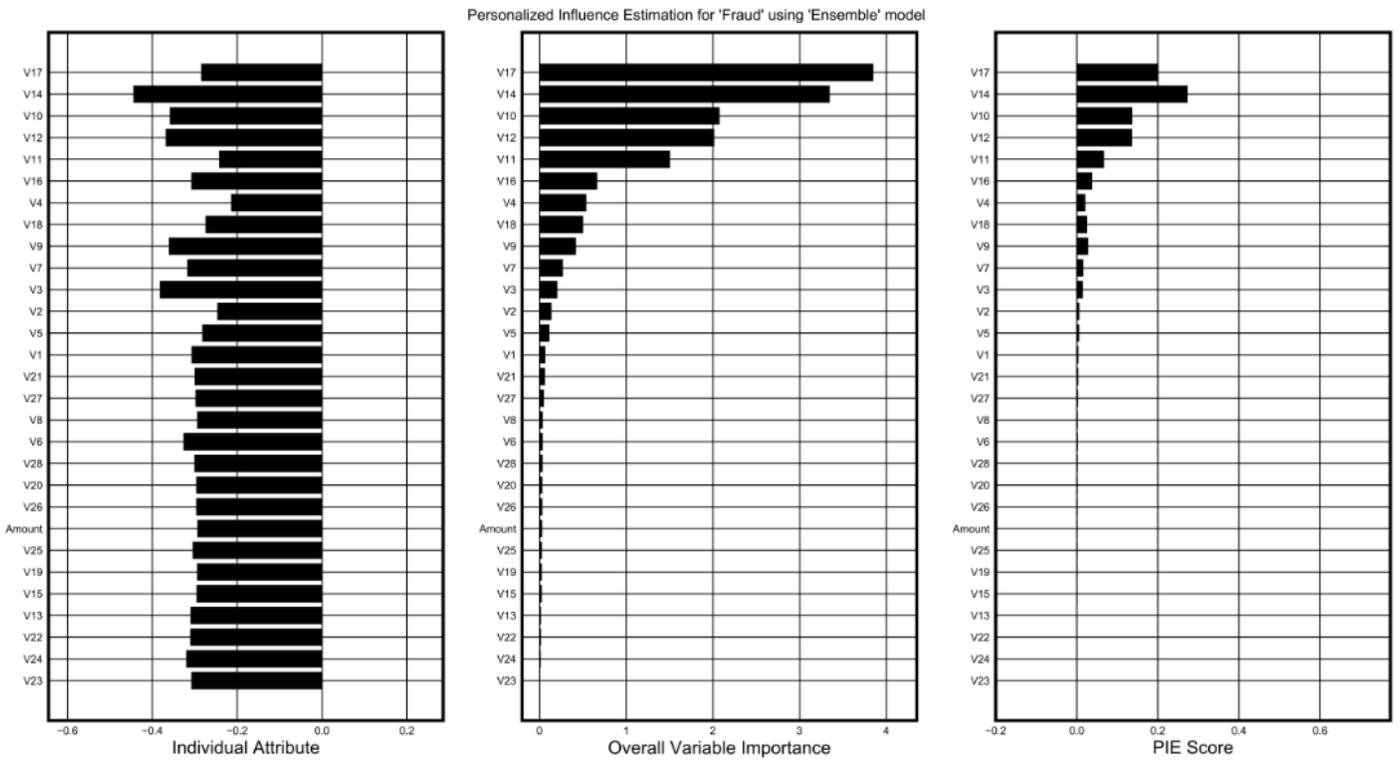

Fig. 5. Personalized Influence Estimation- Tree Model.

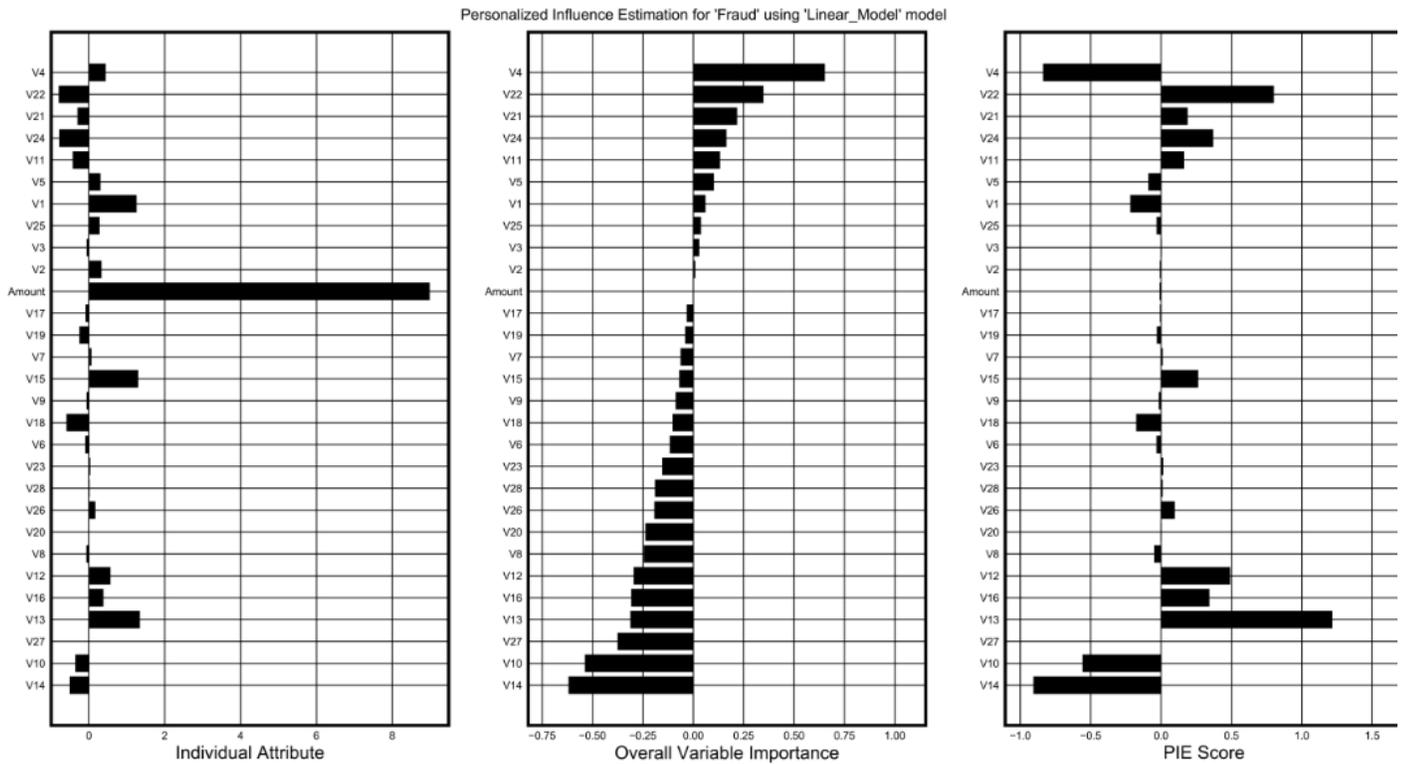

Fig. 6. Personalized Influence Estimation- Linear Model.